\documentclass{article}

\usepackage{arxiv}

\usepackage[utf8]{inputenc} 
\usepackage[T1]{fontenc}    
\usepackage{hyperref}       
\usepackage{url}            
\usepackage{booktabs}       
\usepackage{amsfonts}       
\usepackage{nicefrac}       
\usepackage{microtype}      
\usepackage{lipsum}		
\usepackage{graphicx}
\usepackage{natbib}
\usepackage{color} 
\usepackage{amssymb} 
\usepackage{physics}
\usepackage{empheq}
\usepackage{caption}
\usepackage{subcaption}
\usepackage{breqn}
\usepackage{multirow}
\usepackage{pdfpages}
\usepackage{geometry}
\usepackage{doi}

\usepackage{algorithm}
\usepackage{algpseudocode}

\title{Episodic Memory Theory for the Mechanistic Interpretation of Recurrent Neural Networks}

\author{ Arjun Karuvally \\
  Manning College of Information and Computer Sciences \\
  University of Massachusetts \\
  Amherst, MA \\
  \texttt{akaruvally@umass.edu} \\
   \And
  Peter Delmastro \\
  Department of Mathematics and Statistics \\
  University of Massachusetts, \\
  Amherst, MA \\
  \texttt{pdelmastro@umass.edu}
  \AND
  Hava T. Siegelmann \\
  Manning College of Information and Computer Sciences \\
  University of Massachusetts \\
  Amherst, MA \\
  \texttt{hava@umass.edu}
}

\renewcommand{\shorttitle}{Episodic Memory Theory}

\fancypagestyle{reproducibility}{\fancyhf{}\fancyfoot[L]{Code for reproducing results can be found at: \href{https://github.com/arjunkaruvally/emt_variable_binding}{https://github.com/arjunkaruvally/emt\_variable\_binding}}}
\hypersetup{
pdftitle={Episodic Memory Theory for the Mechanistic Interpretation of Recurrent Neural Networks},
pdfsubject={cs.NE},
pdfauthor={Arjun Karuvally, Peter Delmastro, Hava T. Siegelmann},
pdfkeywords={Recurrent Neural Networks, Mechanistic Interpretability, Memory Models},
}

\begin{document}
\maketitle

\begin{abstract}
	Understanding the intricate operations of Recurrent Neural Networks (RNNs) mechanistically is pivotal for advancing their capabilities and applications. In this pursuit, we propose the Episodic Memory Theory (EMT), illustrating that RNNs can be conceptualized as discrete-time analogs of the recently proposed General Sequential Episodic Memory Model. To substantiate EMT, we introduce a novel set of algorithmic tasks tailored to probe the variable binding behavior in RNNs. Utilizing the EMT, we formulate a mathematically rigorous circuit that facilitates variable binding in these tasks. Our empirical investigations reveal that trained RNNs consistently converge to the variable binding circuit, thus indicating universality in the dynamics of RNNs. Building on these findings, we devise an algorithm to define a \textit{privileged basis}, which reveals hidden neurons instrumental in the temporal storage and composition of variables — a mechanism vital for the successful generalization in these tasks. We show that the privileged basis enhances the interpretability of the learned parameters and hidden states of RNNs. Our work represents a step toward demystifying the internal mechanisms of RNNs and, for computational neuroscience, serves to bridge the gap between artificial neural networks and neural memory models.
\end{abstract}


\section{Introduction}

\thispagestyle{reproducibility}

Mechanistic interpretability aims to reverse engineer the intricate workings of neural networks that drive their behavior \citep{Olah2022}. At the core of its significance lies the pressing need for transparency and comprehension in an era where AI-driven systems have become ubiquitous in real-world applications \citep{Christian2021TheAP,Sears2021ThePE,Bostrom2014SuperintelligencePD,Mller2013FuturePI}. While these systems demonstrate remarkable proficiency, their inherently black-box nature often renders them inscrutable \citep{Alishahi2019AnalyzingAI,Buhrmester2019AnalysisOE,Fong2017InterpretableEO}. Gaining a mechanistic understanding not only builds trust in such systems but also provides insights that can lead to refinement and innovation \citep{Raukur2022TowardTA}. In essence, mechanistic interpretability is not just about demystifying AI; it's about harnessing its potential responsibly and efficiently. 
Current approaches to mechanistic interpretability focus on neural networks without any long-term temporal behavior. 
On the other hand, Recurrent Neural Networks (RNNs) pose a unique challenge - the task relevant information is stored in a hidden state that evolves over time.
This raises the question: 
\textit{How is information reliably stored and processed in an evolving hidden state and how is the hidden state dynamics connected to the computations performed?}
%

To answer this question, we draw inspiration from memory models in computational neuroscience. 
First, we show how autonomously evolving RNNs can be interpreted as performing episodic memory retrieval in Section \ref{section:EpisodicMemoryTheory}. This establishes the connection between existing RNN architectures and neurocomputational memory models, and forms the foundational principle for the Episodic Memory Theory (EMT).
In Section \ref{section:VariableBindingTasks}, we formulate a class of algorithmic tasks to probe the variable binding behavior of RNNs.
Section \ref{section:TheoreticalModelofVariableBinding} introduces \textit{variable memories}, which are linear subspaces capable of symbolically binding and recursively composing information, and use them to reverse engineer variable binding mechanisms in RNNs. 
The experimental results reveal consistent convergence to the proposed mechanisms adding evidence to the \textit{universality hypothesis} in mechanistic interpretability \citep{olah2020zoom, pmlr-v44-li15convergent}. 
In Section \ref{section:VariableMemoryAnalysis}, we build on the empirical results and propose an algorithm to construct a privileged basis grounded in the \textit{variable memories}. For the first time, this basis unveils \textit{hidden neurons}, which exist in superposition within the RNN hidden state, and actively participate in the storage and composition of variables.

\section{Related Works} \label{section:Background}

\begin{figure*}[t]
         \centering
		\includegraphics[scale=0.65]{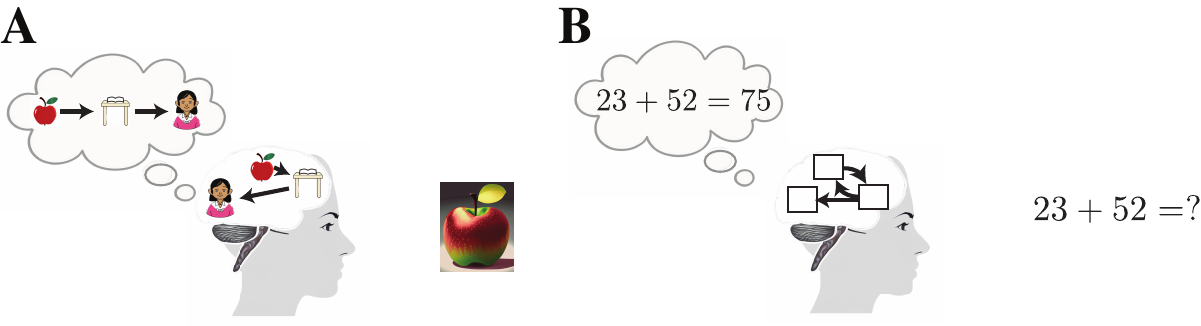}
\caption{ \textbf{Equivalence between Episodic Memory Models and Variable Binding}: \textbf{A}. Episodic Memory models aim to uncover the cognitive processes involved in the retrieval of subjective past experiences often stored as a temporal sequence of memory items. The illustration shows the retrieval of a personal experience when an apple is observed. \textbf{B}. The illustration shows the application of the Episodic Memory Theory, which poses that learning the addition operation over arbitrary numbers, a task involving variable binding, can be considered equivalent to episodic memory retrieval where the computations are performed over variables instead of predetermined memories. The abstract addition operation is stored in the synapses in the form of \textit{how} the variables interact with each other to produce the desired result. }
\label{fig:VB_EM_equivalence}
\end{figure*}

%
%

\textbf{Dynamical Systems Interpretation of RNNs}: Current approaches to interpret  RNNs consider them as non-linear dynamical systems and apply linearization around fixed or slow-changing points to reveal their behavior \citep{Marschall2023ProbingLT,Sussillo2013OpeningTB}. The preliminary step in this analysis involves linearization around fixed points and slow-changing points found using optimization algorithms. The phase space flow is assembled piecemeal from each linearized region. The exploration of the long-term behavior of these regions is undertaken through the eigen-spectrum analysis of the corresponding linearized dynamical systems \citep{Strogatz1994NonlinearDA}, providing insights into the dynamics of convergence, divergence, stability, or spiraling \citep{Rowley2009SpectralAO,Kim1996QUASIPERIODICRA}. However, this method becomes intractable when there are many dimensions exhibiting non-convergent behaviors. The proposed EMT generalizes this approach and enables interpretation even when the number of non-converging dimensions is arbitrarily large. 

\textbf{Mechanistic Interpretability}: Mechanistic interpretability seeks to reverse-engineer neural networks to expose the underlying mechanisms enabling them to learn and adapt to previously unencountered conditions. The prevailing strategy involves examining the networks' internal ``circuits" \citep{Conmy2023TowardsAC,Wang2022InterpretabilityIT,cammarata2020thread:}.
%
%
Researchers have found that applying these interpretability methods to large networks, such as transformers \citep{Vaswani2017AttentionIA} handling complex tasks in natural language processing and vision, faces the challenge of unclear features to be modeled in internal circuits. To address this, they create toy models with clearly defined features essential for task resolution. Probing models trained on toy tasks has resulted in supporting evidence for prevalent hypotheses. Some of the notable hypotheses are \textit{universality} \citep{pmlr-v202-chughtai23a,Li2015ConvergentLD} - models learn similar features and circuits across different models when trained on similar tasks, \textit{bottleneck superposition} \citep{Elhage2022ToyMO} - a mechanism for storing more information than the available dimensions, and \textit{composable linear representations} \citep{Cheung2019SuperpositionOM} - the use of linear spaces in feature representation.
Despite these advancements, current approaches remain confined to networks without a recurrent state like MLPs and transformers. Recurrent architectures, which maintain and recursively update a hidden state for task-related information processing, present a unique challenge - information needs to be stored and processed over time \citep{Cruse1996NeuralNA,Hochreiter1997LongSM}. 

\textbf{Neural Memory Models}: 
%
%
%
Developments in memory modeling have revealed links between deep neural networks and memory models.
The first investigation of this link explored the relationship between Dense Associative Memory and Multi-Layer Perceptron (MLP) with various activation functions \citep{Krotov2016DenseAM}. Later studies extended this connection to explain the practical computational benefits observed in neural architectures like transformers \citep{Ramsauer2020HopfieldNI}. 
%
%
Recently, the traditional memory models capable of associative recall of single memories were expanded to retrieving memory sequences \citep{Karuvally2022EnergybasedGS, Chaudhry2023LongSH}. 
This expansion allows memories that previously did not interact in the single memory retrieval context to interact and produce complex temporal behavior \citep{Kleinfeld1986SequentialSG, Kleinfeld1988AssociativeNN}.
A fundamental assumption in memory modeling (in both single and sequence retrieval) is that the network's memories are predetermined and stored in the synapses. 
This assumption limits the models' applicability to mechanistic interpretability, which requires the symbolic binding of memories typically available only during inference.

In EMT, we will demonstrate that by lifting the fixed memory assumption in memory modeling, these memory models can be utilized to show how binding of external information happens in RNNs, revealing the synergistic relationship between the three fields - memory modeling, recurrent neural networks and mechanistic interpretability. 

\section{RNN as Episodic Memory} \label{section:EpisodicMemoryTheory}

\begin{figure*}[t]
         \centering
		\includegraphics[scale=0.3]{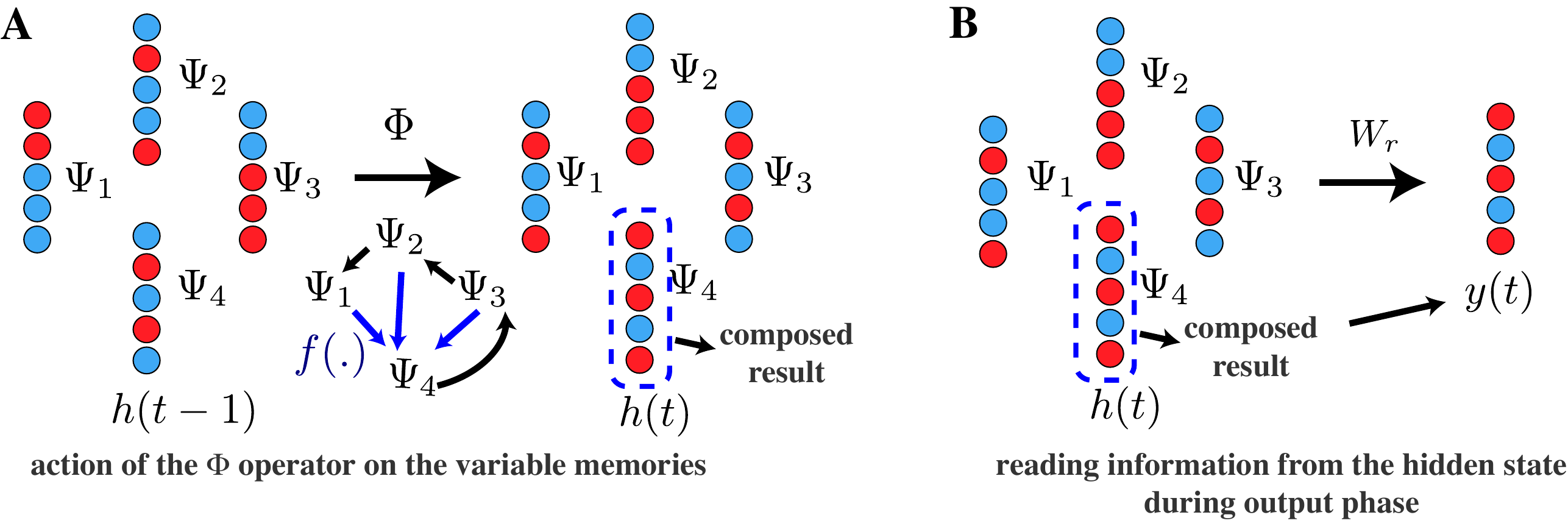}
\caption{ \textbf{Circuit of variable binding in an illustrative task of four variables, each with five dimensions}: \textbf{A}. The hidden state at time $t$ has subspaces capable of storing external information in their activities. The colors depict the vector components (or activity) of the hidden state in the variable memory basis. The linear operator $\Phi$ acts on the hidden state such that these activities are copied between variable memories except for $\Psi_4$, which implements the linear operation $f$. \textbf{B}. The $N^{\text{th}}$ variable contents are read during the output phase using the appropriate linear operator $W_r = \Psi^*_N$. }
\label{fig:VB_Code2RNN}
\end{figure*}

We show that RNNs can be viewed as a discrete-time analog of a memory model called General Sequential Episodic Memory Model (GSEMM) \citep{Karuvally2022EnergybasedGS}.
%
%
To be applicable for the more general setting of RNNs, we slightly modify the GSEMM formulation with a pseudoinverse learning rule instead of the Hebbian learning rule for the synapses. This modification allows us to deal with more general (linearly independent vectors) memories than orthogonal vectors \citep{Chaudhry2023LongSH,Personnaz1986}. 
%
%
We discretize the continuous time GSEMM model using forward Euler approximation under the conditions that the timescale parameters are $\mathcal{T}_f = 1, \mathcal{T}_h = 0,$ and $\mathcal{T}_d = 0$ (See Appendix \ref{appendix:TMVB:RNN-EM-Equivalence} for details).
The final discrete system we obtain is $V_f(t+1) = \Xi \, (I + \Phi^{\prime \top}) \, \Xi^\dag \, \sigma_f(V_f(t))$.
The columns of $\Xi$ are the \textit{stored memories} of the model, and $(I+\Phi^{\prime \top}) = \Phi^\top$ is the matrix representing sequential memory interactions between these stored memories. The neural state variable is $V_f \in \mathbb{R}^{d \times 1}$. 
%
%
The discrete system we derived is topologically conjugate to the update equations of an Elman RNN under the homeomorphic transformation if the norm of the matrix is bounded by 1. That is, if $||\Xi \, \Phi^\top \, \Xi^\dag || \leq 1$, we can consider a new state variable $h = \sigma_f(V_f)$ such that
\begin{equation}
	h(t) = \sigma_f(\Xi \, \Phi^\top \, \Xi^\dag h(t-1)) \,.
        \label{eqn:RNNMemoryDynamics}    
\end{equation}
This conjugate system has equations that are equivalent to an Elman RNN hidden state update equation without bias $h(t+1) = \sigma_f(W_{hh} h(t))$. 

A corollary to the equivalence between the sequence memory model and Elman RNNs is that if we decompose the weight matrix of the RNN in terms of the memories such that $W_{hh} = \Xi \, \Phi^\top \, \Xi^\dag$, the RNN computations can be interpreted as the retrieval of episodic memories temporally transitioning according to the rules encoded in $\Phi$. This result along with the previous results \citep{Krotov2016DenseAM,Ramsauer2020HopfieldNI} connecting memory models with feedforward neural networks forms the foundation of the Episodic Memory Theory of learned neural networks: 
\textbf{The Episodic Memory Theory (EMT) poses that the inner workings of learned neural networks can be revealed by analyzing the learned inter-memory interactions and its effect in the space of stored memories} (Figure \ref{fig:VB_EM_equivalence}).

\section{Variable Binding Tasks} 
\label{section:VariableBindingTasks}

\begin{algorithm}[t]
\caption{Algorithm for computing variable memories of trained linear RNNs}\label{alg:variableMemories}
\begin{algorithmic}
\State $0 \leq \alpha \leq 1$
\State $s$ \Comment{number of time-steps in the input phase}
\State $W_{hh}, W_{uh}, W_{r}$ \Comment{learned parameters of the RNN}
\State $\Psi_s \gets \alpha W_{uh} + (1-\alpha) W_r^*$
\For{$k \in \{ s-1, s-2, \hdots 1 \}$}
     \State $\Psi_k \gets \alpha W^{s-k}_{hh} W_{uh} + \left( 1-\alpha \right) \left(\left( W_{hh}^\top \right)^k W_r^*\right)$
     \State $\Psi_k \gets \Psi_k - E E^* \Psi_k \quad \forall E: \lambda(E) < 1$ \Comment{Remove the components along transient directions.}
\EndFor
\State $\Psi \gets [\Psi_1;\hdots;\Psi_s]$
\State $\Psi^\perp \gets \text{PC}(\{\tilde{h(t)}\} - \Psi^* \, \{\tilde{h}(t)\})$ \Comment{Principle Components of $\tilde{h}$ from simulations}
\end{algorithmic}
\end{algorithm}
Simple algorithmic tasks are very useful for mechanistic interpretability as they provide a controllable empirical setup compared to complex and noisy real-world scenarios.
We formulate a class of tasks with input and output phases. At each timestep of the input phase, external information is provided \textit{to} the RNN. During the output phase, the RNN needs to utilize this external information to synthesize novel outputs at each time step that are subsequently read \textit{from} the network as output. This simple two-phase setup closely matches the behavior of NLP tasks like translation and conditional generative modeling where the noisy real-world features are abstracted out.
Formally, the input phase consists of $s$ total timesteps where at each timestep $t$, a vector of $d$ dimensions $u(t) = \left( u^1(t), u^2(t), \hdots, u^d(t) \right)^\top$ is input to the model. We call the vector components $u^i(t)$ the external information that needs to be \textit{stored} in the RNN hidden state.
After the input phase is complete, the zero vector is continually passed as input to the model, so we say the RNN evolves autonomously (without any external input) during the output phase.
The future states of the system during output phase evolve according to the following equation.
\begin{equation}
    u(t) = f(u(t-1), u(t-2),\hdots u(t-s)), \quad t>s .
    \label{eqn:TaskDynamics}
\end{equation}
For analytical tractability, we add two restrictions to the variable binding tasks: (1) the composition function $f$ is a linear function of its inputs. (2) the codomain of $f$ and the domain of the inputs is binary $\in \{ -1, 1 \}$. 

\section{Variable Binding Circuit in RNN} \label{section:TheoreticalModelofVariableBinding}

Linearization approaches have shown promise in the analysis of RNNs in the neighborhood of fixed points \citep{Sussillo2013OpeningTB}. We build our model of variable binding on a linearized RNN defined by the following equations.
\begin{empheq}{equation}
    \begin{cases}
        h(t) = W_{hh} h(t-1) + W_{u h} u(t) \, , \\
        y(t) = W_r \, h(t) \, .
    \end{cases}
    \label{eqn:ElmanRNNDynamics}
\end{empheq}
We envision that any non-linear RNN can be converted to this form by finding fixed points and subsequent linearization (See Appendix \ref{appendix:TMVB:generalRNNs} for details).
Here, $W_{hh}, W_{uh}, W_r$ are linear operators, $h(t)$ is the hidden state, $u(t)$ is the input, and $y(t)$ is the output. We use a simplifying assumption that $W_{hh}$ has sufficient capacity to represent all the variables required for variable binding tasks (no requirement for bottleneck superposition \citep{Elhage2022ToyMO}). We further assume that $h(0)$ is the zero vector.
To handle the basis change suggested by the EMT view of RNNs, we use Dirac and Einstein notations from abstract algebra (Appendix \ref{appendix:TMVB:MathematicalPreliminaries}).
This new notation has two benefits - (1) We are able to formalize variable binding mechanisms independent of the basis, and (2) The notation enables a very clean and concise description of all the components of the circuit.
Formally, we write any vector $v$ as $\ket{v} = \bra{\epsilon^i} \ket{v} \ket{\epsilon_j} = v^i \ket{\epsilon_j}$. $\ket{\epsilon_j}$ is the basis in which the vector has vector components $v^i = \bra{\epsilon^i}\ket{v}$. $\bra{\epsilon^i}$ are the basis \textit{covectors} defined such that $\bra{\epsilon^i}\ket{\epsilon_j} = \delta_{i j}$, and $\delta_{i j}$ is the Dirac delta function.

In the new notation, the linear RNN of Equation \ref{eqn:ElmanRNNDynamics} evolving autonomously is reformulated as follows.
\begin{equation}
    \ket{h(t)} = \left( \xi^i_{\mu} \, \Phi^\mu_\nu \, (\xi^\dag)^\nu_j \; \ket{e_i} \bra{e^j} \right) \ket{h(t-1)}
\end{equation}
Here $\ket{e_i}$ is the standard basis vector which is typically used in simulations.
To simplify this system further, we use a new basis $\ket{\psi_\mu} = \xi^i_{\mu} \ket{e_i}$.
\begin{dmath}
    \ket{h(t+1)} = \left( \Phi_{\nu}^\mu \, \ket{\psi_\mu} \, \bra{\psi^\nu} \right) \ket{h(t)}
\end{dmath}
This new basis allows us to treat the linearized RNN as applying a \textit{single} linear operator as opposed to three in the original formulation. In the new basis, the hidden state vector is $\ket{h(t)} = h^{\psi_\mu} \ket{\psi_{\mu}}$.
%
%
We pose that these components $h^{\psi_\mu}$ (also called subspace \textit{activity}) can be set to any external information for solving the task by appropriately interacting with the hidden state, thus behaving like variables in computation.
The collection of vectors $\{\Psi_i\} = \{ \ket{\psi_{\mu}} : \mu \in \{ (i-1) d, \hdots, i d \} \} $ that defines a subspace where the variable is stored is called the $i^{\text{th}}$ \textit{variable memory}.
The newly introduced concept of variable memory enables treating activity in the space collectively, simplifying reasoning about their behavior.
In order to retain information, the $\Phi$ operator must have necessary mechanisms to handle the storage of information over time.
One possibility for this is the following linear operator $\Phi$.
\begin{dmath}
    \Phi = \sum_{\mu=1}^{(N-1) d} \ket{\psi_{\mu}} \bra{\psi^{\mu+d}} + \underbrace{\sum_{\mu=(N-1)d}^{N d} \Phi_{\nu}^{\mu} \ket{\psi_\mu} \bra{\psi^\nu}}_{f(u(t-1), u(t-2), \hdots, u(t-N))} \, .
    \label{eqn:PhiTheoretical}
\end{dmath}
The action of the operator on the hidden state is illustrated in Figure \ref{eqn:RNNMemoryDynamics}A. 
For all variable memories with index $i \in \{2, 3, 4, \hdots N\}$, the information contents are copied to the variable memory with index $i-1$. 
The operator then implements the function $f$ defined in Equation \ref{eqn:TaskDynamics} and stores the result in the $N^{\text{th}}$ subspace. Any \textit{linear} function $f$ of the history can be represented in this framework.
This specific setting of the linear operator allows information to be stored in the $N^{\text{th}}$ variable over $N$ timesteps reliably for computing the solution to the problem.
%


\textbf{Reading Variables}: Once RNN has performed its computation, the computed information needs to be extracted. RNNs have a linear operator $W_{r}$, which facilitates the reading of information from $\ket{h(t)}$ at consecutive time steps. We propose that $W_{r}$ has the following equation.
\begin{equation}
    W_{r} = \Psi^*_N = \sum_{\mu=(N-1)d + 1}^{Nd} \ket{e_{\mu - (N-1)d}} \bra{\psi^{\mu}}
\end{equation}
The reading operation reads the activity of the $N^{th}$ subspace and outputs them in the standard basis (Figure \ref{fig:VB_Code2RNN}B) for the output to be read out of RNNs.

We do not propose any form for $W_{uh}$. This is because the operator has two roles to play in the RNN behavior. One role is to add the input information in the $N^{th}$ subspace, and two, suppress the effect of $\Phi$ operator during the input phase -- $\Phi$ actively computes $f$ even though all the variables are not filled during the input phase.
These conflicting roles make proposing a \textit{linear} form for $W_{uh}$ intractable.
In our experiments, we find that this is not an issue for describing the long term behavior of the RNN as $W_{uh}$ only influences the input phase.

\begin{table}[t]
\begin{tabular}{l|lll|lll}
\toprule 
    Task & \multicolumn{3}{c}{hidden size: 64} & \multicolumn{3}{c}{hidden size: 128}\\

      & L2: 0.0 & L2: 0.001 & L2: 0.1 & L2: 0.0 & L2: 0.001 & L2: 0.1 \\
    \midrule
     $\mathcal{T}_1$ & --- (0.97)& --- (0.88)& --- (0.50)& 0.0005  (1.00)& --- (0.93)& --- (0.50)\\
$\mathcal{T}_2$ & 0.0075  (1.00)& --- (0.85)& --- (0.50)& 0.0055  (0.98)& 0.0031  (0.98)& --- (0.50)\\
$\mathcal{T}_3$ & 0.0026  (1.00)& 0.0010  (0.97)& --- (0.50)& 0.0031  (0.98)& 0.0005  (1.00)& --- (0.50)\\
$\mathcal{T}_4$ & 0.0112  (0.94)& 0.0011  (1.00)& --- (0.50)& 0.0022  (1.00)& 0.0006  (1.00)& --- (0.50)\\
\bottomrule
\end{tabular}
\caption{\textbf{RNNs consistently converge to the variable binding model}: The table shows the MAE in the complex argument between the eigenspectrum of the predicted $\Phi$ from the variable binding circuit and the empirically learned $W_{hh}$ in $4$ tasks across $20$ seeds under different RNN configurations. This average error is indeterminate if the rank of the theoretical $\Phi$ is different from the empirical $W_{hh}$. Values in the brackets show the average test accuracy of the trained model. For models that have high test accuracy ($>0.94$), the error in the theoretically predicted spectrum is very low indicating consistent convergence to the theoretical circuit. A notable exception of this behavior is $\mathcal{T}_1$ with hidden size$=64$ and $L2=0$, where the restricted availability of dimensions forces the network to encode variables in bottleneck superposition resulting in a low-rank representation of the solution.}
\label{table:VBConvergence}
\end{table}

\textbf{Optimization}: One point to note is that our definition of variable memory till now is not optimized - there are dimensions in certain variables that are irrelevant for future computations and hence uncessary to be stored. 
This redundant information can be safely discarded without any effect to the network behavior.
Concretely, we propose that RNNs learn an optimized representation of the variables such that for a basis transform $M$ where each standard basis vector is masked by mask $m^i \in \{ 0, 1 \}$. $M = \ket{m_i e_i} \bra{m^j e^j}$:
\begin{equation}
    \arg \min_{m^i} \sum_i m^i \,, \text{such that }, \, rank(M^T \Phi M) = rank(\Phi)
\end{equation}
The final optimized basis is the minimum amount of information that needs to be retained for computing the function $f$ safely over time. For the rest of the paper, we will always consider this optimized basis, unless explicitly specified.

\section{Algorithm for Computing Variable Memories} \label{section:VariableMemoryAnalysis}
The mechanisms we elucidate suggest that \textit{variable memories} can be treated as a privileged and interpretable basis - a special basis where RNN behavior is interpreted as forming variables that are stored and processed. To compute variable memories, we compare the learned hidden weights $W_{hh}$ to the interaction matrix $\Phi$ of the theoretical model in experimental settings. We view the hidden state space in the new basis consisting of the $\Psi = [\Psi_1; ...; \Psi_s]$. We consider also an additional orthogonal basis $\Psi^\perp$ for the remaining dimensions of the space not currently explained by the variable binding circuit in Section \ref{section:TheoreticalModelofVariableBinding}. Once this basis is defined, the action of the learned weights on the variable memories can be extracted from the learned $W_{hh}$ using the basis transformation $\Phi_{\textrm{learned}}^\top = \Psi^* W_{hh} \Psi$. Any interaction between the variable memories and the non-memory space will be encoded in the matrices $\Psi^* W_{hh} \Psi^\perp$ and $(\Psi^\perp)^* W_{hh} \Psi$.

Building on the intuition from the linear model, we use the learned weights $W_{hh},$ and $W_{r}$ to estimate the $\Psi_k$ for the variable memories of the trained RNN. 
The read weights $\Psi_s = W_{r}^*$ define one of the variable memory, and all other subspaces can be found simply by propagating these dimensions \textit{forward} in time: $\Psi_k = \Phi^{s-k} \Psi_s = W_{hh}^{s-k} W_{r}^*$. 
%
%
%
Although the variable memories are defined based on a linear evolution assumption, we found that the method of power iterating $W_{hh}$ was effective in defining a variable memory space for even nonlinear RNNs. 
We specifically selected to define the variable memory dimensions $\Psi_{k}$ by propagating $W_{r}^*$ forward in time: 
\begin{equation}
\Psi_s = W_{r}^*, \quad \quad \Psi_k = W_{hh}^{s-k} W_{r}^*  \quad \textrm{for} \ k < s
\end{equation}
We also removed the projection of each $\Psi_k$ onto the eigenvectors of $W_{hh}$ whose eigenvalues were less than 1 in magnitude since they do not contribute to the long-term behavior but may interfere with the basis definition. 
The theorized interaction matrix $\Phi$ has eigenvalues that sit on the unit circle in the complex plane. This ensures that iterating $\Phi$ does not cause the hidden state to contract or expand in the linear model. We will expect that any eigenvectors of the learned $W_{hh}$ with eigenvalue inside the unit circle corresponding to \textit{transient} behavior associated with dimensions whose activity dies out at later timesteps during inference. 
To obtain the orthogonal basis $\Psi^\perp$ for the rest of the hidden space, we applied principal component analysis (PCA) to hidden state dynamics during inference after removing the projection onto the variable memories.
The algorithm to compute variable memories from a linearized RNN is summarized in Algorithm \ref{alg:variableMemories}.

\section{Results}

\begin{figure*}[t]
         \centering
		\includegraphics[scale=0.67]{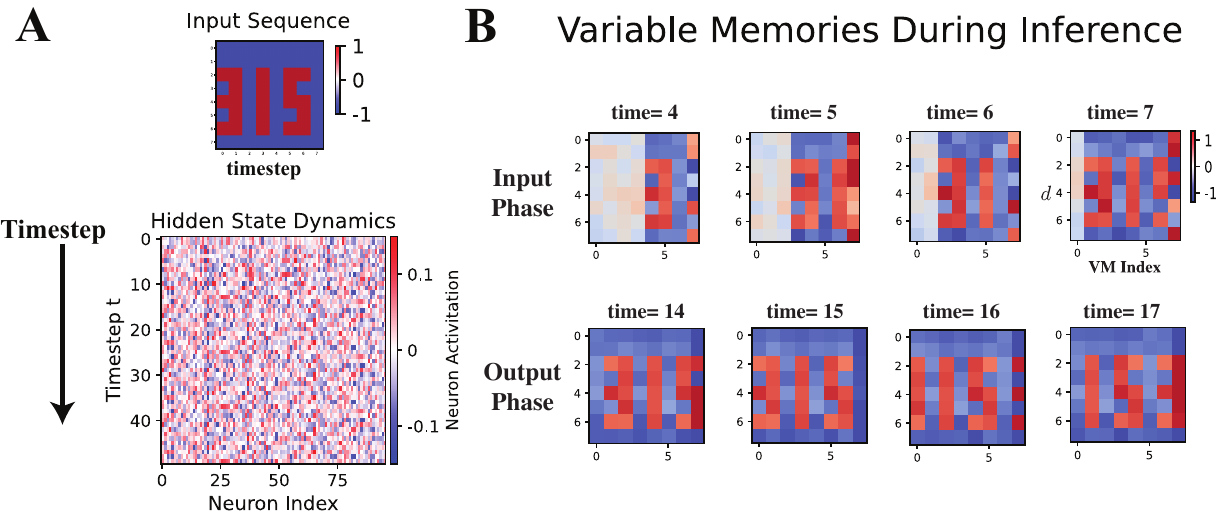}
\caption{ \textbf{EMT reveals hidden neurons storing task relevant information over time}: \textbf{A}. In the repeat copy task ($\mathcal{T}_1$), the RNN needs to repeatedly produce an input sequence that is presented. A typical trained hidden state after providing the input does not show any meaningful patterns connected to the input. \textbf{B}. The same hidden states when visualized in the variable memories reveal the input information being stored as variables and processed according to the variable binding circuit. The actual hidden state is in a superposition of these \textit{hidden} variable memory activities. }
\label{fig:VMHiddenNeurons}
\end{figure*}

\subsection*{Rnns consistently converge to the variable binding model} 

In Section \ref{section:TheoreticalModelofVariableBinding}, we proposed an exact, mathematically rigorous circuit for variable binding capable of storing and processing information over time. To substantiate that this mechanism is indeed learned by RNNs, we trained various RNN configurations, differing in hidden sizes and regularization penalties. After training, the RNNs were linearized, and the eigen-spectrum of the learned $W_{hh}$ matrix is compared with the theoretical $\Phi$, as defined in Equation \ref{eqn:PhiTheoretical}. If RNNs learn a representation in alignment with our model, both operators, i.e., the learned $W_{hh}$ and theoretical $\Phi$, are expected to share a portion of their spectrum as they are similar matrices (i.e they differ by only a basis change). In this comparison, it's pertinent to evaluate solely the arguments of the spectrum, disregarding the magnitude.
The rationale behind this exclusion lies in what the magnitude tells about the dynamical behavior, which portray whether a linear dynamical system is diverging, converging, or maintaining consistency along the eigenvector directions. RNNs typically incorporate a squashing non-linearity, such as the tanh activation function, which restricts trajectories that divergence to infinity.
Essentially, provided the eigenvalue magnitude remains $\geq 1$, the complex argument solely determines the overall dynamical behavior of the RNN. 
Table \ref{table:VBConvergence} depicts the average absolute error when various RNN models are trained across $4$ distinct tasks.
The RNNs exhibiting robust generalization tend to consistently converge towards the circuit mechanisms detailed in Section \ref{section:TheoreticalModelofVariableBinding}.
The results also highlight a particular setup - $\mathcal{T}_1$ with hidden size$=64$ and $L2=0.0$, which, while not converging to the theoretical mechanism, still achieves high generalization accuracy.
$\mathcal{T}_1$ stands out as it necessitates exactly $64$ dimensions per the optimized theoretical model, matching the exact dimensionality available in the RNN.
In this instance, RNNs exhibit a form of bottleneck superposition, a scenario not accommodated by the variable binding circuit yet.
Nevertheless, provided there are sufficient dimensions to encapsulate the variable binding circuit, the RNNs tend to converge to it.

\subsection*{Variable memory reveals hidden neurons storing variable information}

\begin{figure*}[t]
         \centering
		\includegraphics[scale=0.68]{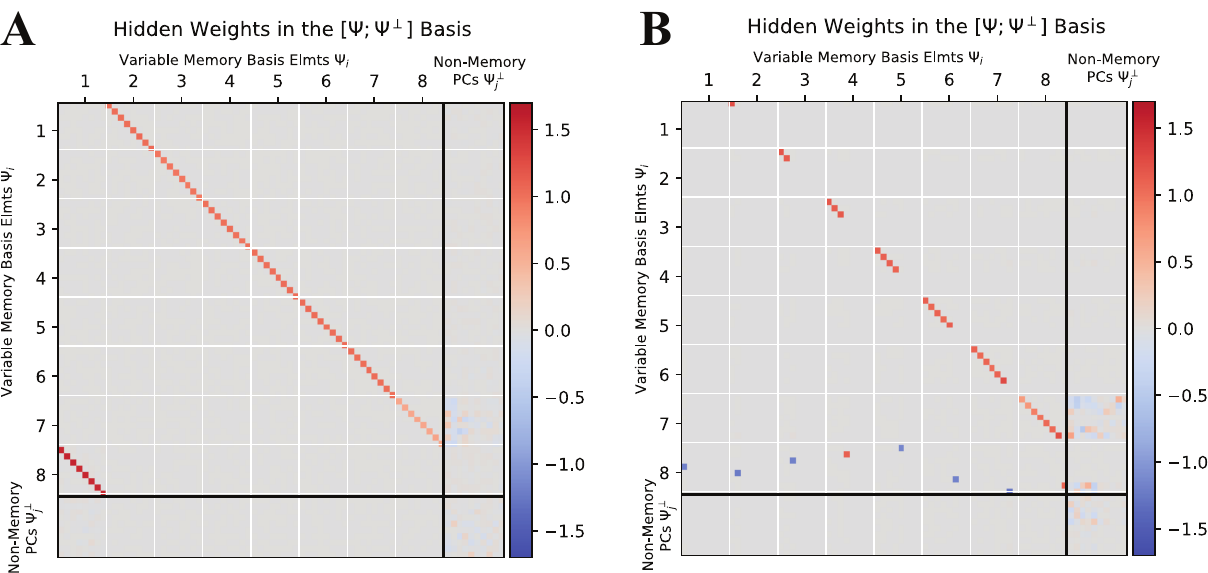}
\caption{ \textbf{EMT enables human interpretability of learned RNN parameters}: The learned weights when visualized in the variable memories result in a form that is human-interpretable. For RNNs trained on two sample tasks $\mathcal{T}_1$ (\textbf{A} left) and $\mathcal{T}_2$ (\textbf{B} right), the weight matrix $W_{hh}$ converts into a form that reveals internal mechanisms of how RNNs solve the task. For both tasks, the variables with index $<8$ copies its contents to the preceding variable. Variable $8$ actively computes the function $f$ applied on all the variables stored in the hidden state using the variable binding circuit. For $\mathcal{T}_1$, it is a simple copy of the $1^{\text{st}}$ variable, and for $\mathcal{T}_2$, it is a linear composition of all the variables Notably, the circuit for $\mathcal{T}_2$ shows an optimized basis where all the irrelevant dimensions are absent. }
\label{fig:VMInterpretability}
\end{figure*}

Variable memories define bases for the storage and processing of variable information within RNNs if the RNNs follow the variable binding circuit.
Building on the empirical results we obtained on the consistent convergence to the variable binding circuit, we compute variable memories using Algorithm \ref{alg:variableMemories} for models trained in Repeat Copy ($\mathcal{T}_1$).
In the Repeat Copy task, the RNN must repeatedly output a stream of inputs provided during the input phase.
The simulated hidden states of learned RNNs are visualized by projecting the hidden state in the variable memories: $\tilde{h} = \Psi^* h$.
The results shown in Figure \ref{fig:VMHiddenNeurons} reveal that the hidden state is in a superposition of hidden neurons that actively store each variable required to compute the function $f$ at all points in time.
The basis transformation helps to disentangle these superposed variables from the hidden state so that they are easily visualized.

\subsection*{Variable memories enable human interpretability of learned weights}

In addition to revealing hidden neurons that store and process information over time, variable memories can also be used as bases to view the operations of the learned matrices.
The variable memories are carefully constructed such that $W_{hh}$ converts to the underlying $\Phi$ when viewed in the basis.
As observed in the Figure \ref{fig:VMInterpretability}, viewing the operations of $W_{hh}$ in the new basis enables human-interpretability in terms of the variable binding circuit and provides a way to influence or ``fix" RNN behavior after training.
This ``fixing" operation can be imagined as changing specific weights of the extracted $\Phi$ to improve either variable storage properties, or problems computing $f$.

One practical consideration when computing variable memories is the sensitivity of these results to minor changes in basis definition. 
It may be found that in some practical cases, the basis transformation does not reveal these interpretable representations even though the spectrum converges to the theoretical circuit.
This deviation from the theory is a result of the sensitivity of the basis definition to minor errors in the pseudo-inverse required to compute the dual.
It is possible that new empirical procedures may be developed to improve this computation in the future.

\section{Discussion}
 We presented a novel perspective on Recurrent Neural Networks (RNNs), framing them as dynamical systems performing sequence memory retrieval. We introduced the concept of ``variable memories," linear subspaces capable of symbolically binding and recursively composing information. Our approach addresses the limitations of current methods in understanding RNNs, particularly their inscrutability as `black boxes' and the complexity of spectral analysis in high-dimensional task spaces. 
 We presented a new class of algorithmic tasks that are designed to probe the variable binding behavior of RNNs.
 We presented a circuit mechanism that is capable of recursively storing and composing variables and show that trained RNNs consistently converge to this circuit pointing towards universality in the learned models.
 Building on the empirical evidence, we used variable memories to derive a privileged basis that, for the first time, revealed hidden neurons actively involved in information processing in RNNs.
Further, using variable memories, we viewed the learned parameters of an RNN in a human-interpretable manner, enabling reasoning about RNN behavior as repeatedly copying and composing variables.

The Episodic Memory Theory and variable memories are versatile enough to be broadly applicable in various scenarios, offering valuable insights for researchers designing new algorithms. One practical application of this analysis can be in the development of continual learning algorithms, which can restrict gradients to pre-existing variable memory spaces to minimize catastrophic forgetting of prior tasks. Additionally, in task composition—where a new task is a combination of two existing tasks—the linear spaces of each task can be linearly combined to efficiently solve the composite problem. Another possible application is transfer learning, where task knowledge is shared between networks. The Episodic Memory Theory suggests that variable memories and their interactions are the essential components for knowledge transfer, allowing the remaining network dynamics to be disregarded or easily relearned, streamlining the transfer process.

With these diverse applications, it is also important to recognize certain inherent limitations to the approach. One of the limitation is that the analysis is primarily restricted to linear dynamical systems.  Although an accurate representation of the qualitative behavior within small neighborhoods of fixed points can be found for non-linear dynamical systems, the RNNs have to be confined to these linear regions for the approach to be applicable. 
It is an interesting behavior that models consistently converge to these linearized regime, at least for the tasks outlined in Section \ref{section:VariableBindingTasks}. 
The second limitation of the variable binding model is that external information is stored as a \textit{linear} superposition of variable memories in the hidden state. Our results indicates that the role of non-linearity in encoding external information may be minimal for the toy tasks. However, we have observed that when the number of dimensions of the linear operator $W_{hh}$ is not substantially large compared to the task's dimensionality requirements (bottleneck superposition) or when the regularization penalty is high, the RNN can effectively resort to non-linear encoding mechanisms to store external information (Appendix \ref{appendix:ERVB:FurtherExamples}).
Overcoming these limitations of non-linearity will be an interesting direction to pursue in future research.

\bibliographystyle{unsrtnat}
\bibliography{references}  

\appendix

\section{Theoretical Models of Variable Binding} \label{appendix:section:TheoreticalModelsOfVariableBinding}

\subsection{Mathematical Preliminaries} \label{appendix:TMVB:MathematicalPreliminaries}
The core concept of the episodic memory theory is basis change, the appropriate setting of the stored memories. Current notations lack the ability to adequately capture the nuances of basis change.
Hence, we introduce abstract algebra notations typically used in theoretical physics literature to formally explain the variable binding mechanisms. We treat a \textit{vector} as an abstract mathematical object invariant to basis changes. 
Vectors have \textit{vector components} that are associated with the respective basis under consideration. We use Dirac notations to represent vector $v$ as - $\ket{v} = \sum_i v^i \ket{e_i}$.
Here, the linearly independent collection of vectors $\ket{e_i}$ is the \textit{basis} with respect to which the vector $\ket{v}$ has component $v^i \in \mathbb{R}$. Linear algebra states that a collection of basis vectors $\ket{e_i}$ has an associated collection of \textit{basis covectors} $\bra{e^i}$ defined such that $\bra{e^i} \ket{e_j} = \delta_{i j}$,
where $\delta_{i j}$ is the Kronecker delta. This allows us to reformulate the vector components in terms of the vector itself as $\ket{v} = \sum_{i} \bra{e^i} \ket{v} \ket{e_i}$.
We use the Einstein summation convention to omit the summation symbols wherever the summation is clear. Therefore, vector $\ket{v}$ written in basis $\ket{e_i}$ is
\begin{dmath}
    \ket{v} = v^i \ket{e_i} = \bra{e^i} \ket{v} \ket{e_i} .
\end{dmath}
The set of all possible vectors $\ket{v}$ is a \textit{vector space} spanned by the basis vectors $\ket{e_i}$. A \textit{subspace} of this space is a vector space that is spanned by a subset of basis vectors $\{ \ket{e^\prime_j}:  \ket{e^\prime_j} \subseteq \{ \ket{e_i} \} \}$.

The RNN dynamics presented in Equation \ref{eqn:RNNMemoryDynamics} represented in the new notation is reformulated as: 
\begin{dmath} \label{eq:elmanRNN}
    \ket{h(t)} = \sigma_f \left( \left( \xi^i_{\mu} \, \Phi^\mu_\nu \, (\xi^\dag)^\nu_j \; \ket{e_i} \bra{e^j} \right) \ket{h(t-1)} \right) = \sigma_f \left( W_{hh} \, \vec{h}(t-1) \right) .
\end{dmath}
The greek indices iterate over memory space dimensions $\{ 1, 2, \hdots, N_h \}$, alpha numeric indices iterate over feature dimension indices $\{ 1, 2, \hdots, N_f \}$.
Typically, we use the standard basis in our simulations. For the rest of the paper, the standard basis will be represented by the collection of vectors $\ket{e_i}$ and the covectors $\bra{e^i}$. The hidden state at time $t$ in the standard basis is denoted as $\ket{h(t)} = \bra{e^j} \ket{h(t)} \ket{e_i}$.
$\bra{e^j} \ket{h(t)}$ are the \textit{vector components} of $\ket{h(t)}$ we obtain from simulations.
\subsection{RNN-Episodic Memory Equivalence} \label{appendix:TMVB:RNN-EM-Equivalence}
We modify the GSEMM formulation with a pseudoinverse learning rule instead of the Hebbian learning rule for the synapses. This modification allows us to deal with more general (linearly independent vectors) memories than orthogonal vectors \cite{Personnaz1986}. The dynamical equations for our modified GSEMM are given below.
\begin{empheq}{equation}
\begin{dcases}
	\mathcal{T}_f \dv{V_f}{t} =& \sqrt{\alpha_s} \, \Xi \, \sigma_h(V_h) - V_f , \\
	\mathcal{T}_h \dv{V_h}{t} =& \sqrt{\alpha_s} \, \Xi^\dag \, \sigma_f(V_f) + \alpha_c \Phi^{\prime \top} \Xi^\dag V_{d} - V_h , \\
	\mathcal{T}_d \dv{V_d}{t} =& \sigma_f(V_f) - V_d ,
\end{dcases}
\end{empheq}
The neural state variables of the dynamical system are $V_f \in \mathbb{R}^{N_f \times 1}, V_h \in \mathbb{R}^{N_h \times 1}, V_d \in \mathbb{R}^{N_f \times 1}$. The interactions are represented by $\Xi \in \mathbb{R}^{N_f \times N_h}$ and $\Phi \in \mathbb{R}^{N_h \times N_h}$.
$\Xi^\dag$ is the Moore-Penrose pseudoinverse of $\Xi$.
To derive the connection between the continuous time model and discrete updates of RNNs, we discretize the continuous time model using forward Euler approximation under the conditions that $\mathcal{T}_f = 1, \mathcal{T}_h = 0,$ and $\mathcal{T}_d = 0$.
From a given time $t$, the update equations are given as

\begin{empheq}{equation}
\begin{dcases}
	\mathcal{T}_f (V_f(t+1) - V_f(t)) =& \Xi \, \sigma_h(V_h(t)) - V_f(t) \, , \\
	V_h(t) =& \Xi^\top \, \sigma_f(V_f(t)) + \Phi^\top \Xi^\top V_{d}(t) \, , \\
	V_d(t) =& \sigma_f(V_f(t)) \, .
\end{dcases}
\end{empheq}
\begin{empheq}{equation}
\begin{dcases}
	\mathcal{T}_f (V_f(t+1) - V_f(t)) =& \Xi \, \sigma_h(V_h) - V_f(t) \, , \\
	V_h(t) =& \Xi^\top \, \sigma_f(V_f(t)) + \Phi^\top \Xi^\top \sigma_f(V_f) \, , \\
\end{dcases}
\end{empheq}
\begin{empheq}{equation}
\begin{dcases}
	\mathcal{T}_f (V_f(t+1) - V_f(t)) =& \Xi \, V_h - V_f(t) \, , \\
	V_h(t) =& (I + \Phi^\top) \Xi^\top \sigma_f(V_f) \, , \\
\end{dcases}
\end{empheq}
\begin{dmath}
	\mathcal{T}_f (V_f(t+1) - V_f(t)) = \Xi \, (I + \Phi^\top) \Xi^\top \sigma_f(V_f) - V_f(t)
\end{dmath}
Final discrete upate equation
\begin{dmath}
	V_f(t+1) = \Xi \, (I + \Phi^\top) \Xi^\top \sigma_f(V_f)
	\label{governingDynamics:original}
\end{dmath}
Restrict the norm of matrix $||\Xi \, (I + \Phi^\top) \Xi^\top|| \leq 1$. \\
This allows us to consider the transformation $V'_f = \sigma_f(V_f)$, so for invertible $\sigma_f$, 
\begin{dmath}
	\sigma^{-1}_f(V'_f(t+1)) = \Xi \, (I + \Phi^\top) \Xi^\top V'_f
\end{dmath}
\begin{dmath}
	\sigma^{-1}_f(V'_f(t+1)) = \Xi \, (I + \Phi^\top) \Xi^\top V'_f
\end{dmath}
\begin{dmath}
	V'_f(t+1) = \sigma_f(\Xi \, (I + \Phi^\top) \Xi^\top V'_f)
	\label{governingDynamics:rnn}
\end{dmath}
this is a general update equation for an RNN without bias. The physical interpretation of this equation is that the columns of $\Xi$ stores the individual \textit{memories} of the system and the linear operator $(I+\Phi)$ is the temporal interaction between the stored \textit{memories}.
In the memory modeling literature, it is typical to consider memories as a fixed collection instead of a variable collection that shares a common interaction behavior. We will show how in the next sections how the dynamics as a result of fixed collection can be used to store variable information.

\textbf{Topological Conjugacy with RNNs}:
Proof that dynamical systems governed by Equations \ref{governingDynamics:original} and \ref{governingDynamics:rnn} are topological conjugates.

Consider $f(x) = \Xi \, (I + \Phi^\top) \Xi^\top \sigma_f(x)$ for Equation \ref{governingDynamics:original} and $g(x) = \sigma_f(\Xi \, (I + \Phi^\top) \Xi^\top x)$ for Equation \ref{governingDynamics:rnn}. Consider a homeomorphism $h(y) = \sigma_f(y)$ on $g$. Then,
\begin{dmath}
	(h^{-1} \circ g \circ h) (x) = \sigma_f^{-1}( \sigma_f(\Xi \, (I + \Phi^\top) \Xi^\top \sigma_f(x)) ) = \Xi \, (I + \Phi^\top) \Xi^\top \sigma_f(x) = f(x)
\end{dmath}
So, for the homeomorphism $h$ on $g$, we get that $h^{-1} \circ g \circ h = f$ proving that $f$ and $g$ are topological conjugates. Therefore all dynamical properties of $f$ and $g$ are shared.

\subsection{Example: Repeat Copy ($\mathcal{T}_1$)} \label{appendix:TMVB:example}

Repeat Copy is a task typically used to evaluate the memory storage characteristics of RNNs since the task has a deterministic evolution represented by a simple algorithm that stores all input vectors in memory for later retrieval.
Although elementary, repeat copy provides a simple framework to work out the variable binding circuit we theorized in action.
For the repeat copy task, the linear operators of the RNN has the following equations.
\begin{empheq}{equation}
    \begin{cases}
        \Phi = \sum_{\mu=1}^{(s-1) \kappa} \ket{\psi_{\mu}} \bra{\psi^{\mu+\kappa}} + \sum_{\mu=(s-1)\kappa + 1}^{s \kappa} \ket{\psi_\mu} \bra{\psi^{\mu - (s-1) \kappa}} \\
        W_{uh} = \Psi_{s} \\
        W_r = \Psi^*_{s}
    \end{cases}
\end{empheq}
This $\phi$ can be imagined as copying the contents of the subspaces in a cyclic fashion. That is, the content of the $i^{th}$ subspace goes to $(i-1)^{\text{th}}$ subspace with the first subspace being copied to the $N^{\text{th}}$ subspace.
%
The dynamical evolution of the RNN is represented at the time step $1$ as,
\begin{dmath}
    \ket{h(1)} = \ket{\psi_{(s-1)\kappa + j}} \bra{e^j} u^i(1) \ket{e_i}
\end{dmath}
\begin{dmath}
    \ket{h(1)} = u^i(1) \ket{\psi_{(s-1)\kappa + j}} \bra{e^j} \ket{e_i}
\end{dmath}
\begin{dmath}
    \ket{h(1)} = u^i(1) \ket{\psi_{(s-1)\kappa + j}} \delta_{i j}
\end{dmath}
Kronecker delta index cancellation
\begin{dmath}
    \ket{h(1)} = u^i(1) \ket{\psi_{(s-1)\kappa + i}}
\end{dmath}
At time step $2$,
\begin{dmath}
    \ket{h(2)} = u^i(1) \, \Phi \, \ket{\psi_{(s-1)\kappa + i}} + u^i(2) \, \ket{\psi_{(s-1)\kappa + i}}
\end{dmath}
Expanding $\Phi$
\begin{dmath}    
    \ket{h(2)} = u^i(1) \, \left( \sum_{\mu=1}^{(s-1) \kappa} \ket{\psi_{\mu}} \bra{\psi^{\mu+\kappa}} + \sum_{\mu=(s-1)\kappa + 1}^{s \kappa} \ket{\psi_\mu} \bra{\psi^{\mu - (s-1) \kappa}} \right) \, \ket{\psi_{(s-1)\kappa + i}} + u^i(2) \, \ket{\psi_{(s-1)\kappa + i}}
\end{dmath}
\begin{dmath}    
    \ket{h(2)} = u^i(1) \, \ket{\psi_{(s-2)\kappa + i}} + u^i(2) \, \ket{\psi_{(s-1)\kappa + i}}
\end{dmath}
At the final step of the input phase when $t=s$, $\ket{h(s)}$ is defined as:
\begin{dmath}    
    \ket{h(s)} = \sum_{\mu=1}^s u^i(\mu) \, \ket{\psi_{(\mu-1)\kappa + i}}
\end{dmath}
For $t$ timesteps after $s$, the general equation for $\ket{h(s+t)}$ is:
\begin{dmath}    
    \ket{h(s+t)} = \sum_{\mu=1}^s u^i(\mu) \, \ket{\psi_{\left[ \left(\left(\mu-t-1 \mod s\right) + 1 \right)\kappa + i \right]}}
\end{dmath}
From this equation for the hidden state vector, it can be easily seen that the $\mu^{\text{th}}$ variable is stored in the $\left[(\mu-t-1 \mod s) + 1\right]^{\text{th}}$ subspace at time step $t$. The readout weights $W_r = \Psi_s^*$ reads out the contents of the $s^{\text{th}}$ subspace.

\subsection{Application to General RNNs} \label{appendix:TMVB:generalRNNs}

The linear RNNs we discussed are powerful in terms of the content of variables that can be stored and reliably retrieved. The variable contents, $u^i$, can be any real number and this information can be reliably retrieved in the end using the appropriate readout weights.
However, learning such a system is difficult using gradient descent procedures. To see this, setting the components of $\Phi$ to anything other than unity might result in dynamics that is eventually converging or diverging resulting in a loss of information in these variables.
Additionally, linear systems are not used in the practical design of RNNs. The main difference is now the presence of the nonlinearity. In this case, our theory can still be used. 
To illustrate this, consider a general RNN evolving according to $h(t+1) = g(W_{hh} h(t) + b)$ where $b$ is a bias term. Suppose $h(t) = h^*$ is a fixed point of the system. We can then linearize the system around the fixed point to obtain the linearized dynamics in a small region around the fixed point. 
\begin{dmath}
    h(t+1) - h^* = \mathcal{J}(g)|_{h^*} \, W_{hh} \, (h(t+1) - h^*) + O((h(t+1) - h^*)^2)
\end{dmath}
where $\mathcal{J}$ is the jacobian of the activation function $g$. If the RNN had an additional input, this can also be incorporated into the linearized system by treating the external input as a control variable
\begin{dmath}
    h(t+1) - h^* = \mathcal{J}(g)|_{h^*} \, W_{hh} \, (h(t) - h^*) + \mathcal{J}(g)|_{h^*} \, W_{uh} u(t)
\end{dmath}
Substituting $h(t) - h^* = h^{\prime}(t)$
\begin{dmath}
    h^{\prime}(t+1) = \mathcal{J}(g)|_{h^*} \, W_{hh} \, h^{\prime}(t) + \mathcal{J}(g)|_{h^*} \, W_{uh} u(t)
\end{dmath}
which is exactly the linear system which we studied where instead of $W_{hh} = \Xi \Phi \Xi^\dag$, we have $J(g)|_{h^*} W_{hh} = \Xi \Phi \Xi^\dag$. 
%

\section{Experiments}

\subsection{Data} \label{appendix:ERVB:Data}

We train RNNs on the variable binding tasks described in the main paper with the following restrictions - the domain of $u$ at each timestep is binary $\in \{-1, 1\}$ and the function $f$ is a linear function of its inputs. We collect various trajectories of the system evolving accoding to $f$ by first sampling uniformly randomly, the input vectors.
The system is then allowed to evolve with the recurrent function $f$ over the time horizon defined by the training algorithm.

\subsection{Training Details} \label{appendix:ERVB:TrainingDetails}

\textbf{Architecture} \hspace{0.25cm} 
We used single layer Elman-style RNNs for all the variable binding tasks. Given an input sequence $(u(1), u(2), ..., u(T))$ with each $u(t)\in \mathbb{R}^d$, an Elman RNN produces the output sequence $y = (y(1),...,y(T))$ with $y(t) \in \mathbb{R}^{N_{out}}$ following the equations
\begin{equation}
h(t+1) = \tanh(W_{hh} h(t) + W_{uh} u(t)) \quad,\quad y(t) = W_r h(t)
\end{equation}
Here $W_{uh} \in \mathbb{R}^{N_h \times N_{in}}$, $W_{hh} \in \mathbb{R}^{N_h \times N_h}$, and $W_r \in \mathbb{R}^{N_{out} \times N_h}$ are the input, hidden, and readout weights, respectively, and $N_h$ is the dimension of the hidden state space. 
%

The initial hidden state $h(0)$ for each model was \textit{not} a trained parameter; instead, these vectors were simply generated for each model and fixed throughout training. We used the zero vector for all the models.

\vspace{0.25cm}
\noindent \textbf{Task Dimensions} \hspace{0.25cm} Our results presented in the main paper for the repeat copy $(\mathcal{T}_1)$ and compose copy ($\mathcal{T}_2$) used vectors of dimension $d = 8$ and sequences of $s = 8$ vectors to be input to the model.

\vspace{0.25cm}
\noindent \textbf{Training Parameters} \hspace{0.25cm}
We used PyTorch's implementation of these RNN models and trained them using Adam optimization with MSE loss. We performed a hyperparameter search for the best parameters for each task --- see table \ref{table:training-params} for a list of our parameters for each task. 

\begin{table}[h!]
\begin{center}
\begin{tabular}{| c | c | c |}
\hline
 & Repeat Copy ($\mathcal{T}_1$) & Compose Copy ($\mathcal{T}_2$) \\ \hline
 Input \& output dimensions & 8 & 8 \\ \hline
 Input phase (\# of timesteps) & 8 & 8 \\ \hline
 training horizon & 100 & 100 \\ \hline
Hidden dimension $N_h$ & 128 & 128 \\ \hline
 \# of training iterations & 45000 & 45000 \\ \hline
 (Mini)batch size & 64 & 64  \\ \hline
 Learning rate & $10^{-3}$ & $10^{-3}$  \\ \hline
 applied at iteration & 36000 & \\ \hline
 Weight decay ($L^2$ regularization) & none & none \\ \hline
 Gradient clipping threshold & $1.0$ & 1.0 \\ \hline
\end{tabular}
\caption{Architecture, Task, \& Training Parameters}
\label{table:training-params}
\end{center}
\end{table}
\noindent \textbf{Curriculum Time Horizon} \hspace{0.25cm} 
When training a network, we adaptively adjusted the number of timesteps after the input phase during which the RNN's output was evaluated. We refer to this window as the \textit{training horizon} for the model. 

Specifically, during training we kept a rolling average of the model's \textit{loss by timestep} $L(t)$, i.e. the accuracy of the model's predictions on the $t$-th timestep after the input phase. This metric was computed from the network's loss on each batch, so tracking $L(t)$ required minimal extra computation. 

The network was initially trained at time horizon $H_0$ and we adapted this horizon on each training iteration based on the model's loss by timestep. Letting $H_n$ denote the time horizon used on training step $n$, the horizon was increased by a factor of $\gamma = 1.2$ (e.g. $H_{n+1} \gets \gamma H_n$) whenever the model's accuracy $L(t)$ for $t \le H_{\textrm{min}}$ decreased below a threshold $\epsilon = 3 \cdot 10^{-2}$. Similarly, the horizon was reduced by a factor of $\gamma$ is the model's loss was above the threshold ($H_{n+1} \gets H_n / \gamma$). We also restricted $H_n$ to be within a minimum training horizon $H_0$ and maximum horizon $H_{\textrm{max}}$. These where set to 10/100 for the repeat copy task and 10/100 for the compose copy task.

We found this algorithm did not affect the results presented in this paper, but it did improve training efficiency, allowing us to run the experiments for more seeds.

\subsection{Repeat Copy: Further Examples of Hidden Weights Decomposition} \label{appendix:ERVB:FurtherExamples}

This section includes additional examples of the hidden weights decomposition applied to networks trained on the repeat copy task. 

\begin{figure*}[h]
         \centering
		\includegraphics[width=0.73\textwidth]{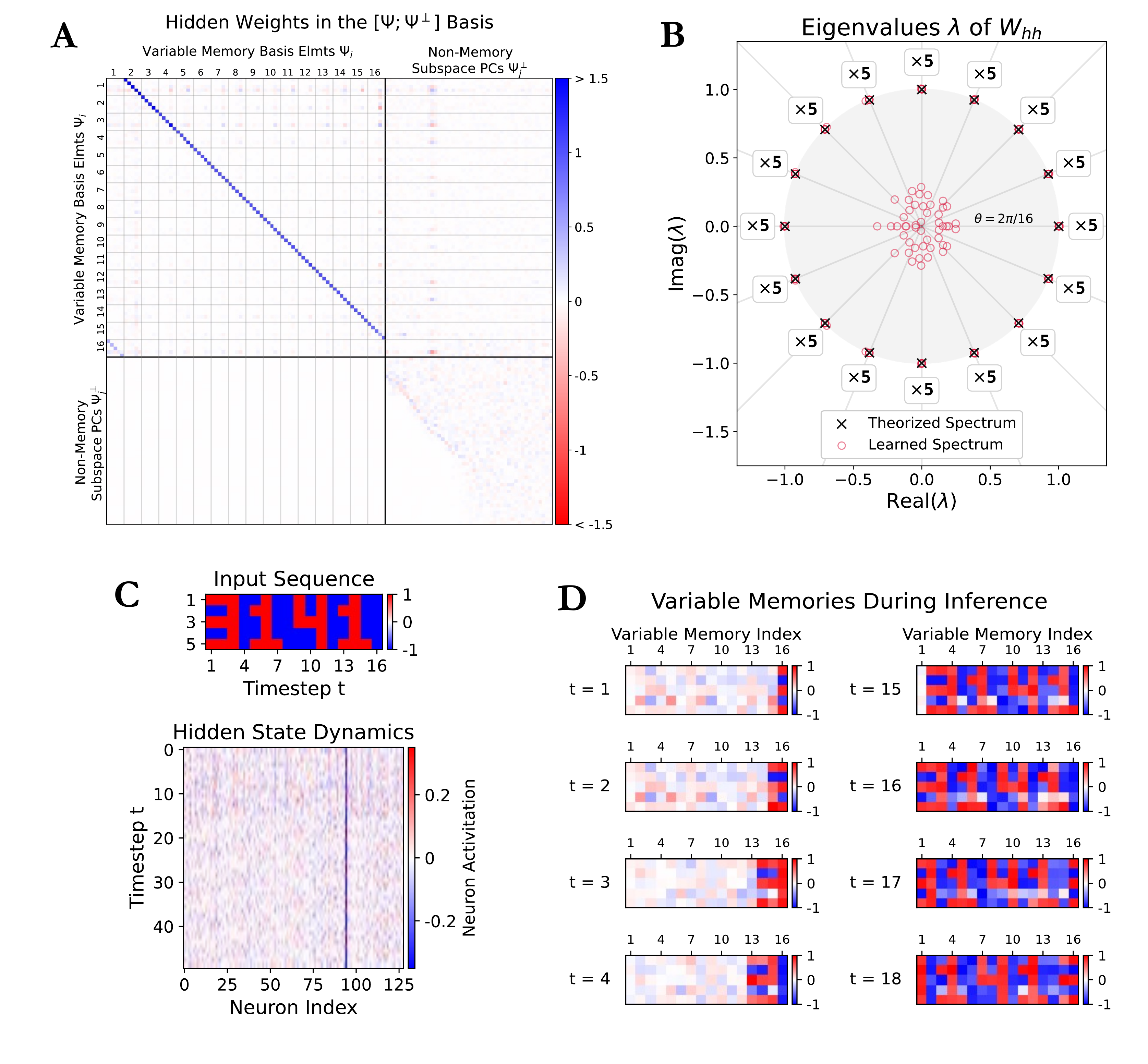}
\caption{ \textbf{Additional Experimental Results of Repeat Copy Task with 16 vectors, each of 5 dimensions}: \textbf{A}. $W_{hh}$ visualized in the variable memory basis reveals the variable memories and their interactions. \textbf{B}. After training, the eigenspectrum of $W_{hh}$ with a magnitude $\geq 1$ overlaps with the theoretical $\Phi$. The boxes show the number of eigenvalues in each direction.
\textbf{C}. During inference, "3141" is inserted into the network in the form of binary vectors. This input results in the hidden state evolving in the standard basis, as shown. How this hidden state evolution is related to the computations cannot be interpreted easily in this basis.
\textbf{D}. When projected on the variable memories, the hidden state evolution reveals the contents of the variables over time. Note that in order to make these visualization clear, we needed to normalize the activity along each variable memory to have standard deviation 1 when assigning colors to the pixels. The standard deviation of the memory subspaces varies due to variance in the strength of some variable memory interactions. These differences in interaction strengths does not impede the model's performance, however, likely due to the nonlinearity of the activation function. Unlike the linear model, interaction strengths well above 1 cannot cause hidden state space to expand indefinitely because the tanh nonlinearity restricts the network's state to $[-1,1]^{N_h}$. This property appears to allow the RNN to sustain stable periodic cycles for a range of interaction strengths above 1.}
\label{fig:RC-S16-d5-example2}
\end{figure*}

\begin{figure*}[h]
         \centering
		\includegraphics[width=0.9\textwidth]{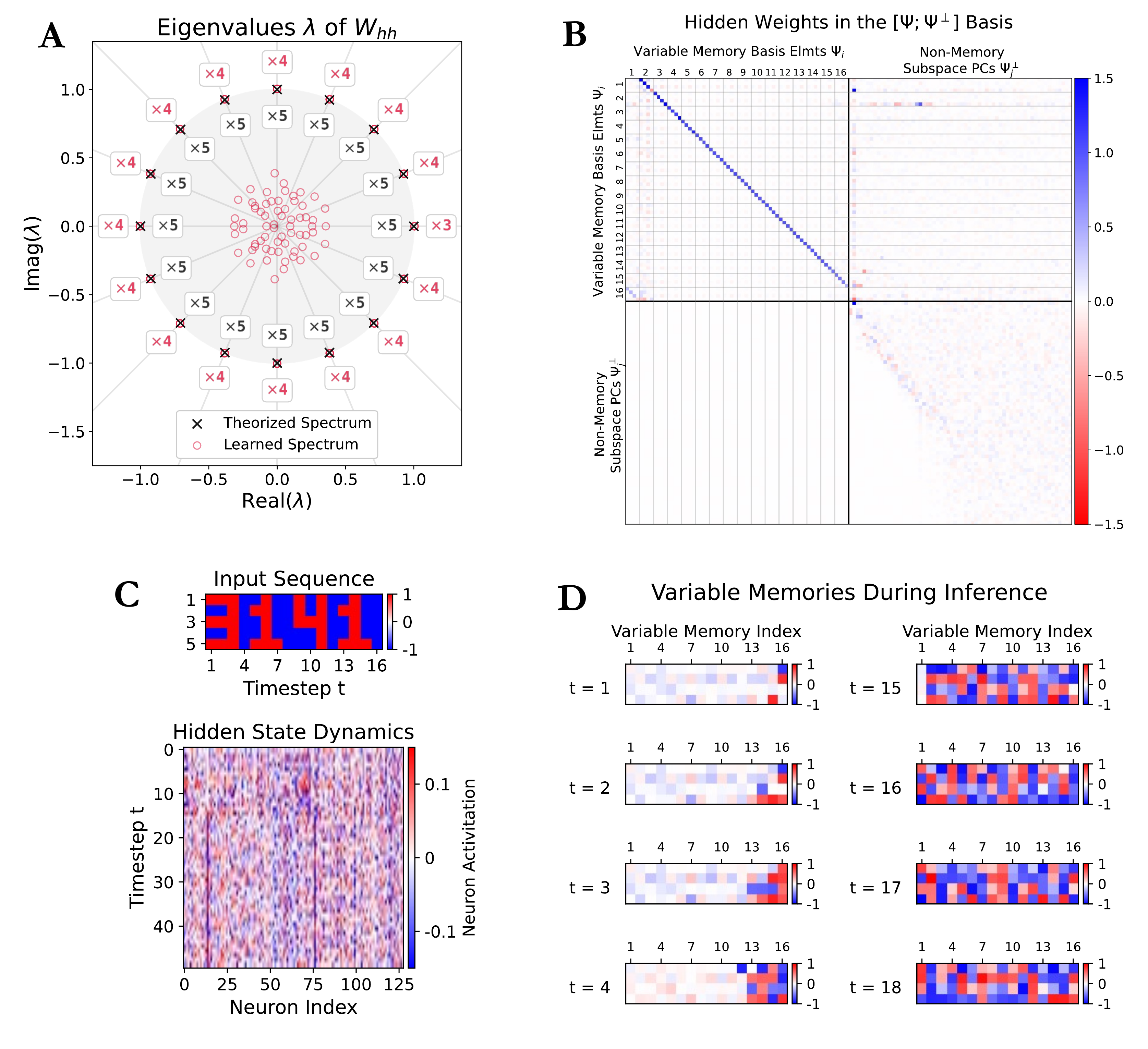}
\caption{ \textbf{Nonlinear Variable Memories Learned for the Repeat Copy Task with 16 vectors, each of 5 dimensions}. \textbf{A}. Eigenspectrum of $W_{hh}$ after training. The learned eigenvalues cluster into 16 groups equally spaced around the unit circle, but there are only 3-4 eigenvectors per cluster (indicated in red). Compare this to the theorized linear model, which has 5 eigenvalues per cluster (indicated in black). The task requires $16 \cdot = 5 = 80$ bits of information to be stored. Linearization about the origin predicts that the long-term behavior of the model is dictated by the eigenvectors with eigenvalue outside the unit circle because its activity along other dimensions will decay over time. The model has only $16 \cdot \mathbf{4} - 1 = 63$ eigenvectors with eigenvalue near the unit circle, so this results suggests the model has learned a non-linear encoding that compresses 80 bits of information into 63 dimensions. 
\textbf{B}: $W_{hh}$ visualized in the variable memory basis reveals the variable memories and their interactions. Here, the variable memories have only 4 dimensions because the network has learned only 63 eigenvectors with eigenvalue near the unit circle. The variable memory subspaces also have non-trivial interaction with a few of the the non-memory subspaces.
\textbf{C}. During inference, "3141" is inserted into the network in the form of binary vectors. This input results in the hidden state evolving in the standard basis, as shown. How this hidden state evolution is related to the computations cannot be interpreted easily in this basis.
\textbf{D}. When projected on the variable memories, the hidden state evolution is still not easily interpreted for this network, likely due to a nonlinear variable memories. As in the previous figure, we normalized the activity along each variable memory to have standard deviation 1 when assigning colors to the pixels.}
\label{fig:RC-S16-d5-nonlinear}
\end{figure*}

\begin{figure*}[h]
         \centering
		\includegraphics[width=0.9\textwidth]{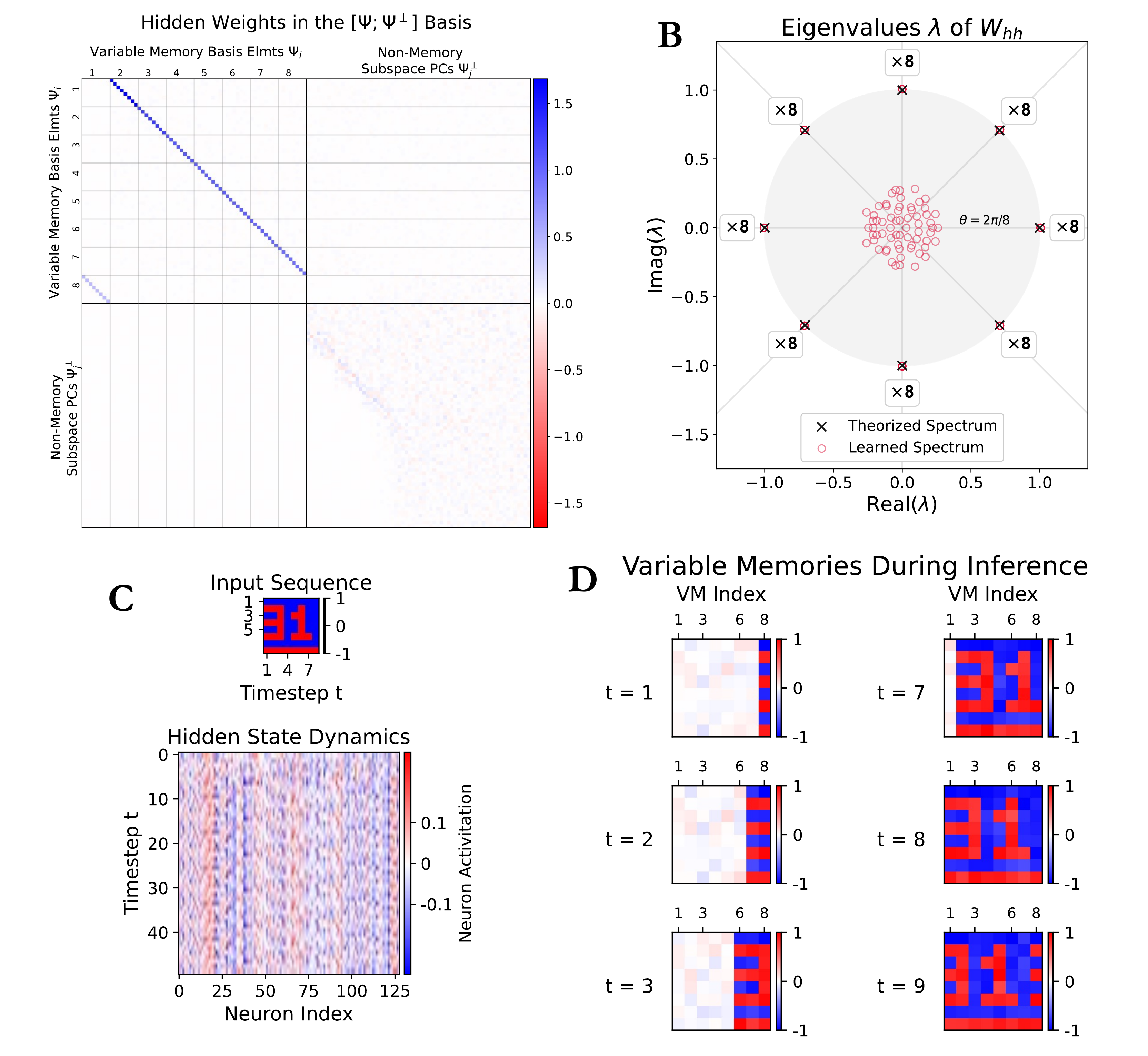}
\caption{ \textbf{Additional Experimental Results of Repeat Copy Task with 8 vectors, each of 8 dimensions}. This figure was included to show the decomposition applied to other values of $s$ and $d$ for the Repeat Copy task. \textbf{A}: $W_{hh}$ visualized in the variable memory basis reveals the variable memories and their interactions. \textbf{B}. After training, the eigenspectrum of $W_{hh}$ with a magnitude $\geq 1$ overlaps with the theoretical $\Phi$. The boxes show the number of eigenvalues in each direction.
\textbf{C}. During inference, "3141" is inserted into the network in the form of binary vectors. This input results in the hidden state evolving in the standard basis, as shown. How this hidden state evolution is related to the computations cannot be interpreted easily in this basis.
\textbf{D}. When projected on the variable memories, the hidden state evolution reveals the contents of the variables over time. As in the previous figures, we normalized the activity along each variable memory to have standard deviation 1 when assigning colors to the pixels.}
\label{fig:RC-S8-d8-example2}
\end{figure*}

\clearpage
\subsection{Uniform vs. Gaussian Parameter Initialization}

\begin{figure*}[h]
\centering
\begin{subfigure}{0.46\textwidth}
    \centering
    \includegraphics[width=\textwidth]{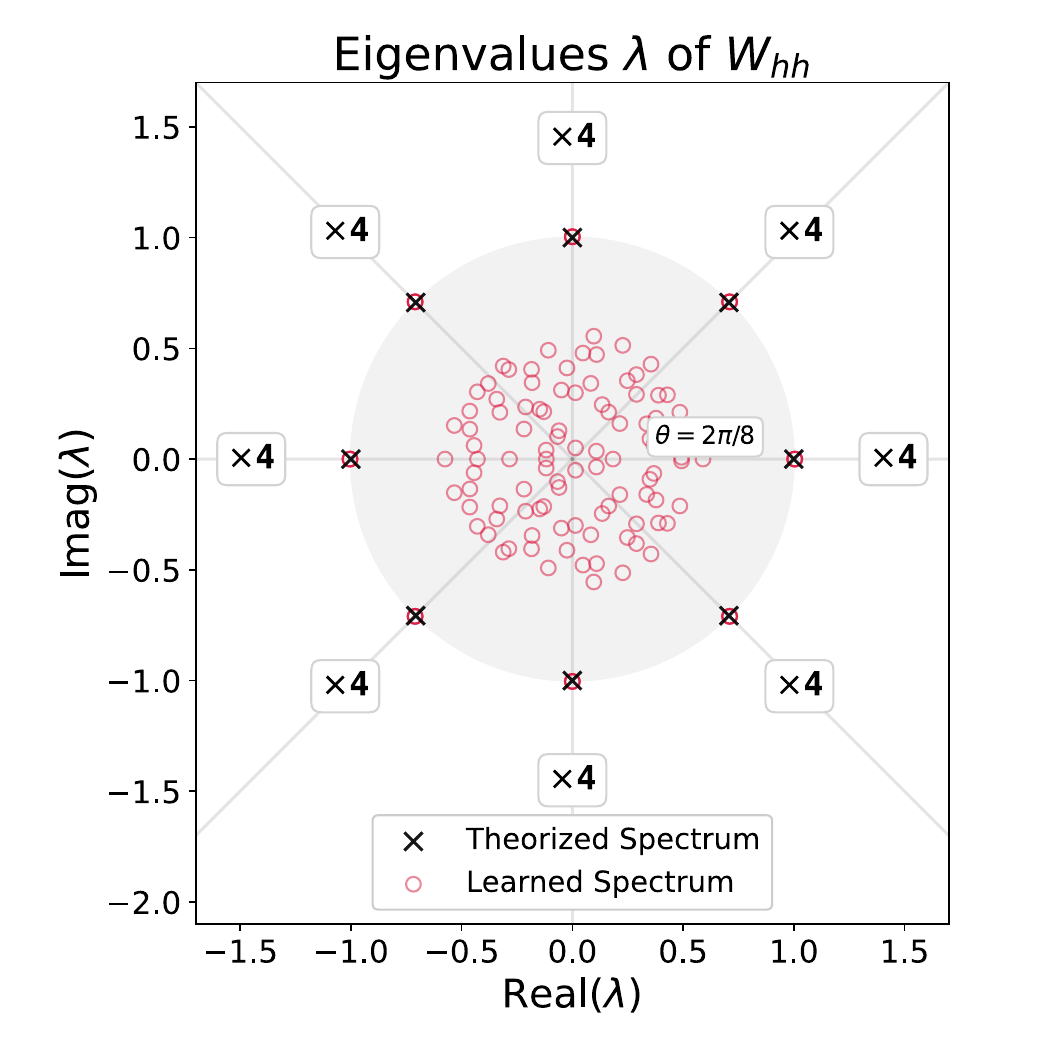}
    \caption{Uniform Initialization}
\end{subfigure}
\begin{subfigure}{0.46\textwidth}
    \centering
    \includegraphics[width=\textwidth]{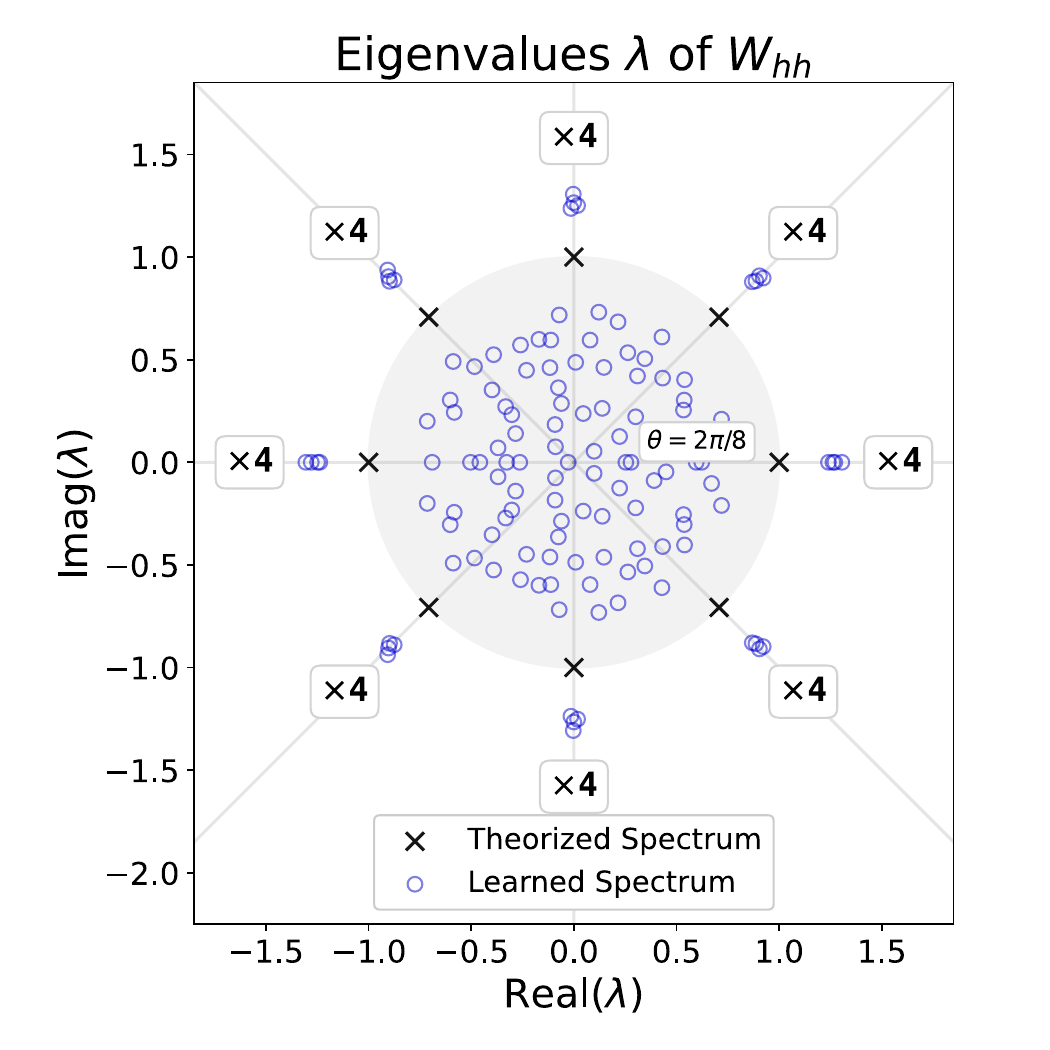}
    \caption{Gaussian Initialization}
\end{subfigure}
\caption{ \textbf{An effect of parameter initialization for the Repeat Copy Task} with $s = 8$ vectors, each of dimension $d = 4$. \textbf{A}: Spectrum (in red) of the learned hidden weights $W_{hh}$ for a network whose parameters where initialized from a uniform distribution over $[-k,k]$ with $k = 1/\sqrt{N_h}$. This network has 32 eigenvalues that are nearly on the unit circle. These eigenvalues are clustered into groups of 4, each group being an angle of $\theta = 2\pi/s$ apart from each other. These eigenvalues coincide with the eigenvalues of the linear model for solving the repeat copy task. \textbf{B}: Spectrum (in blue) of the learned hidden weights $W_{hh}$ for a network whose parameters where initialized from a Gaussian distribution with mean 0 and variance $1/N_h$. This network has 32 eigenvalues outside the unit circle, but they are a larger radii than the model initialized using the uniform distribution. These eigenvalues still cluster into eight groups of four, and the average complex argument of each group aligns with the complex arguments of the eigenvalues for the linear model.
}
\label{fig:RC_gaussian}
\end{figure*}

We also tested a different initialization scheme for the parameters $W_{uh}, W_{hh}$, and $W_{hy}$ of the RNNs to observe the effect(s) this would have on the structure of the learned weights. The results presented in the main paper and in earlier sections of the Supplemental Material used PyTorch's default initialization scheme: each weight is drawn \textit{uniformly} from $[-k,k]$ with $k = 1/\sqrt{N_h}$. Fig. \ref{fig:RC_gaussian} shows the resulting spectrum of a trained model when it's parameters where drawn from a Gaussian distribution with mean 0 and variance $1/N_h$. One can see that this model learned a spectrum similar to that presented in the main paper, but the largest eigenvalues are further away from the unit circle. This result was observed for most seeds for networks trained on  the repeat copy task with $s = 8$ vectors of dimension $\kappa = 4$ and $\kappa = 8$, though it doesn't hold for every seed. We also find that the networks whose spectrum has larger eigenvalues usually generalize longer in time than the networks with eigenvalues closer to the unit circle.

\end{document}